\documentclass{article}
\usepackage{spconf,amsmath,amssymb,graphicx,hyperref,multirow}

\title{MSCT : DIFFERENTIAL CROSS-MODAL ATTENTION FOR DEEPFAKE DETECTION}
\name{Fangda Wei$^{1}$ \qquad Miao Liu$^{1}$ \qquad Yingxue Wang$^{2}$ \qquad Jing Wang$^{1}$ \qquad Shenghui Zhao$^{1 \star}$ \qquad Nan Li$^{2}$\thanks{Thanks to National Natural Science Foundation of China(No.62571037) and Science and Technology Project of Inner Mongolia Autonomous Region (2023YFSW0006) agency for funding.}}
\address{$^{1}$ Beijing Institute of Technology, China\\
      $^{2}$ China Academy of Electronics and Information Technology, China}
\begin{document}
%
\maketitle
\begin{abstract}
Audio-visual deepfake detection typically employs a complementary multi-modal model to check the forgery traces in the video. These methods primarily extract forgery traces through audio-visual alignment, which results from the inconsistency between audio and video modalities. However, the traditional multi-modal forgery detection method has the problem of insufficient feature extraction and modal alignment deviation. To address this, we propose a multi-scale cross-modal transformer encoder (MSCT) for deepfake detection. Our approach includes a multi-scale self-attention to integrate the features of adjacent embeddings and a differential cross-modal attention to fuse multi-modal features. Our experiments demonstrate competitive performance on the FakeAVCeleb dataset, validating the effectiveness of the proposed structure.
\end{abstract}
\section{Introduction}
In recent years, deep generative algorithms have advanced rapidly. Notably, the rapid development of technologies like variational autoencoders (VAE) \cite{a1}, generative adversarial networks (GAN) \cite{a2}, and diffusion models \cite{i1} has enabled easy creation of synthetic videos---posing severe threats to individuals, society, and nations. To address deepfakes, researchers have proposed solutions spanning single modalities (video, audio) and multi-modality.

Single-modal deepfake detection relies on a single input modality, such as audio \cite{i7} or vision \cite{i8}. However, these methods are constrained by their single-modality input, making them hard to handle more flexible deepfake generation techniques. To address this limitation, numerous studies have proposed multi-modal deepfake detection algorithms \cite{i2} that fuse information from audio and video modalities.
\begin{figure}[htb]
\begin{minipage}[b]{1.0\linewidth}
  \centerline{\includegraphics[width=8cm]{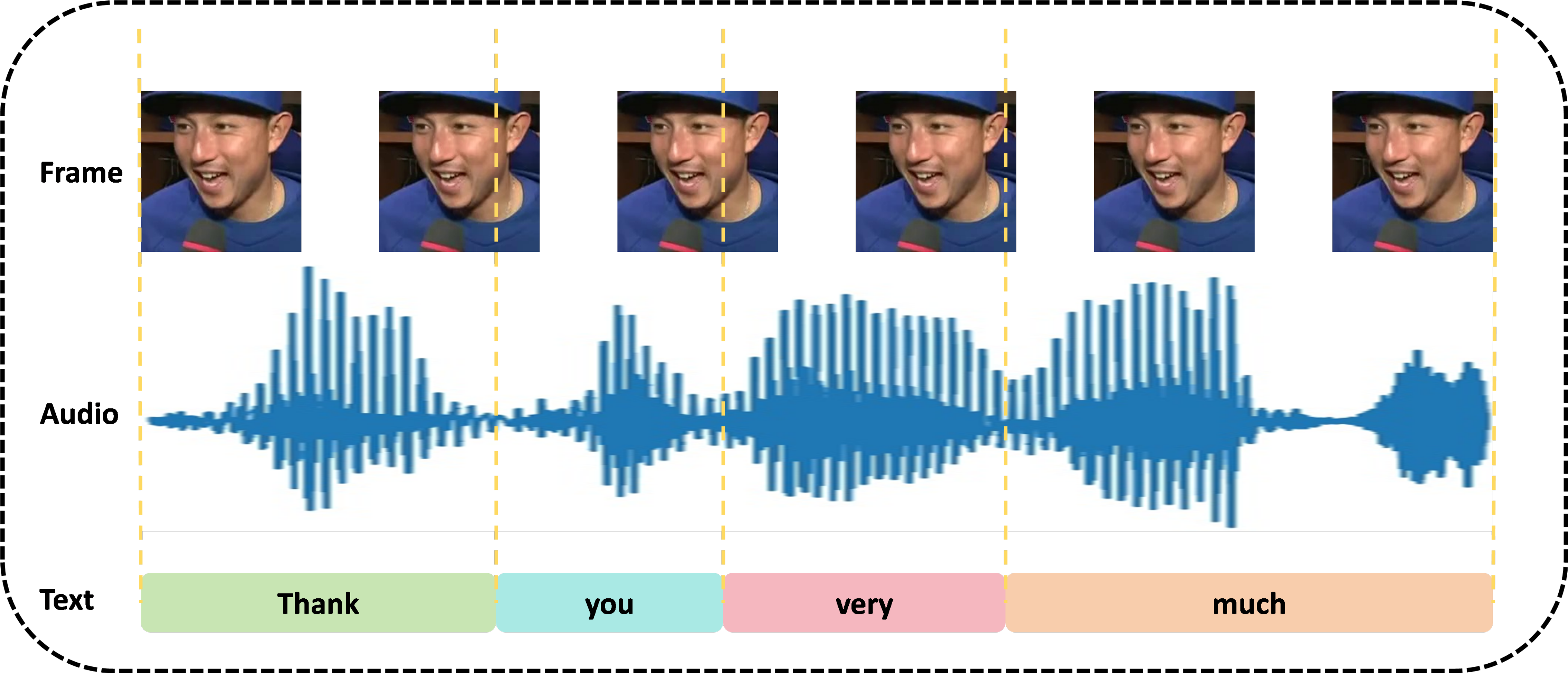}}
\end{minipage}
\caption{The video and audio parts corresponding to each frame contains incomplete text information}
\label{img1}
\vspace{-0.5cm}
\end{figure}
Current multi-modal deepfake detection methods primarily leverage audio-visual consistency, focusing on inter-modal inconsistencies between audio and video. For instance, Mittal et al. \cite{i9} used audio-video emotional mismatch as a cue for multi-modal deepfake detection. Chugh et al. \cite{i10} employed a modal detuning score (MDS) to quantify audio-video discrepancies. Zou et al. \cite{i2} proposed intra-modal and inter-modal regularization techniques, enhancing multi-modal model performance via audio-visual consistency.

As noted earlier, most multi-modal deepfake detection models enhance performance by aligning modalities via cross-modal similarity measurement. With the advancement of attention mechanisms, some works \cite{a3,i12} have improved alignment by fusing audio-video features through cross-modal attention. However, we argue that traditional cross-modal attention may conflict with multi-modal forgery detection. Specifically, cross-modal attention derives queries and keys from different modalities and generates attention matrices via matrix multiplication---yet most deepfake detection tasks rely on modal alignment losses for cross-modal alignment. These losses require cross-modal similarity of fake videos to approach 0 and that of real videos to approach 1. As a result, attention matrices of fake videos are constrained while those of real videos are enhanced, reducing the model's sensitivity to forged video regions but increasing it to real content. Such real-video-biased attention is detrimental to deepfake detection.
\begin{figure*}[htb]
\begin{minipage}[b]{1.0\linewidth}
  \centerline{\includegraphics[width=0.8\textwidth]{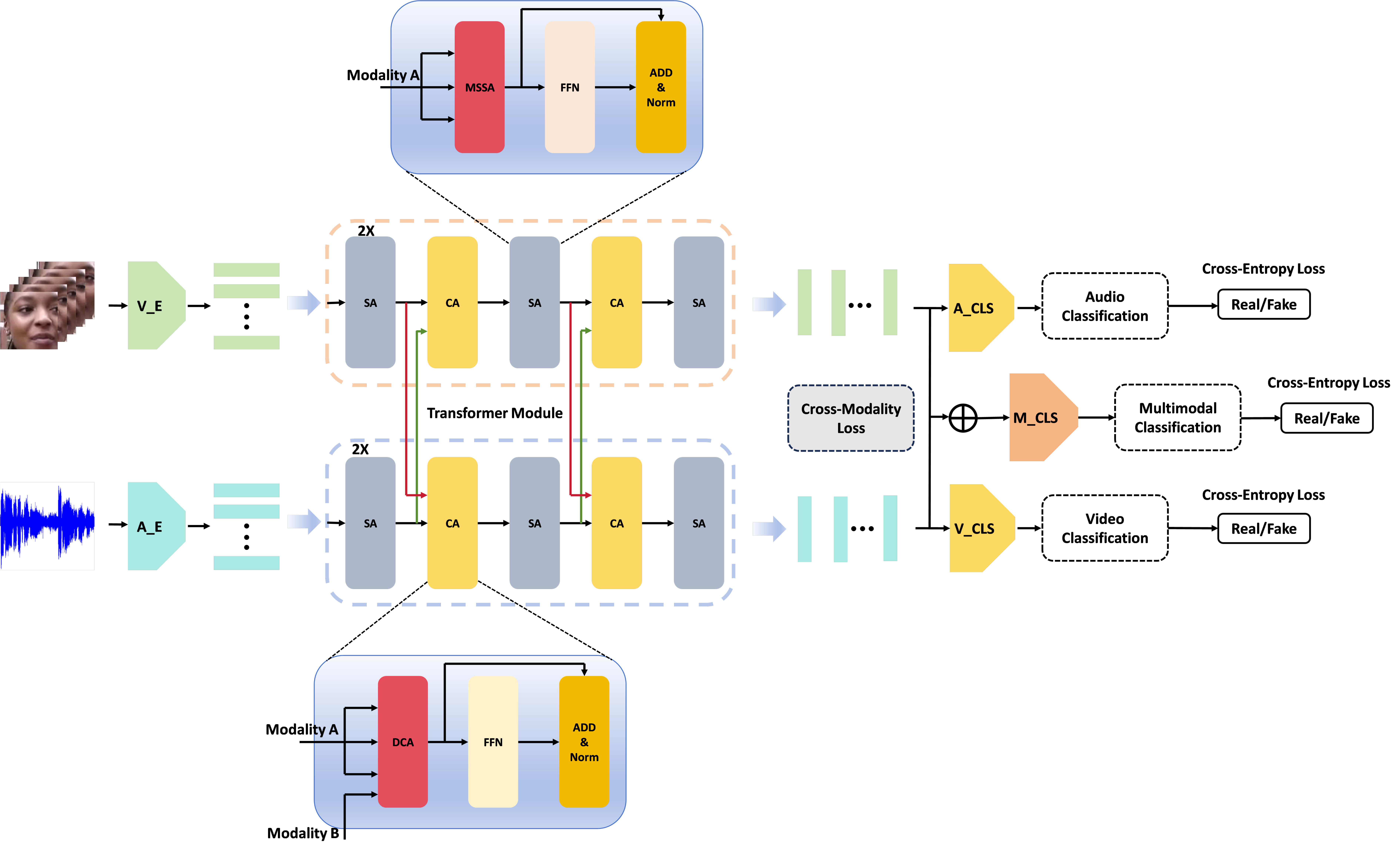}}
\end{minipage}
\caption{Our proposed approach consists of encoder and classification block (A\_CLS, V\_CLS, M\_CLS). The encoder contains pre-encoder (A\_E, V\_E) and transformer module, which consists of self attention (SA) and cross attention (CA).}
\label{img2}
\vspace{-0.5cm}
\end{figure*}
Additionally, current models demand stronger temporal awareness. Most algorithms only extract complex spatial information within frames \cite{i2}, while lacking multi-scale temporal feature extraction. In most cases, a single frame contains incomplete information---its remaining semantic details may reside in adjacent frames---yet frame-level feature extraction fails to fully capture comprehensive semantic information, as illustrated in \textbf{Fig.\ref{img1}}.

To address the problems, we propose two task-specific attention modules for more accurate extraction of forgery information. First, we design a differential cross-modal attention module: by introducing attention matrix differences, it enables the model to better focus on fake video cues and enhances compatibility between cross-modal attention and cross-modal alignment loss. Second, we propose a multi-scale self-attention module: it extracts multi-scale features in the temporal dimension, allowing each embedding to adaptively integrate information from adjacent embeddings. Evaluated on the FakeAVCeleb dataset, our method achieves outstanding performance.
\section{METHODOLOGY}
This section introduces our multi-modal deepfake detection framework (\textbf{Fig.\ref{img2}}), which comprises a single-modal feature extraction module and a multi-modal feature fusion module. Built on this framework, we focus on detailing our proposed multi-scale self-attention module and differential cross-modal attention module.
\subsection{Audio-visual deepfake detection model}
The pre-processed audio and visual channel inputs are denoted as $x_{a} \in \mathbb{R}^{B \times C_{a} \times T}$ and $x_{v} \in \mathbb{R}^{B \times C_{v} \times T \times H \times W}$. The $C_{a}=104$ and $C_{v}=3$ represent the number of audio and visual feature channels, respectively. The overall multi-modal detection labels are denoted as $y_{m}$. In addition, $y_{a}$ and $y_{v}$ are defined as the individual labels for the audio and visual modalities.
\subsubsection{Feature extraction and regularization}
Firstly, we obtain the output $f_{i} \in \mathbb{R}^{B \times T \times C}$ of each modality via the pre-encoder, where $i$ represents the specific modality and $C$ represents the feature dimension. To further extract the forgery features, $f_{i}$ is fed into the transformer module, yielding the output $Z_{i}=[z_{cls}, z_{1}, z_{2}, ..., z_{T}]$ with $z_{t} \in \mathbb{R}^{B \times C}$

Following MRDF-CE \cite{i2}, We regularize the transformer output $[z_{1},..., z_{T}]$. Specifically, we use cross-modal alignment loss to align paired audio-visual signals, and cross-entropy-based modality-specific regularization to preserve modality-specific details. For multi-modal classification, we concatenate the $z_{cls}$ of each modality and feed them into the classifier.
\subsection{Attention module}
To better align alignment loss with cross-modal attention, we propose a differential cross-modal attention module. Furthermore, to equip the transformer encoder with the capability of multi-scale time information perception, we introduce a multi-scale self-attention module.
\subsubsection{Differential cross-modal attention module}
\begin{figure}[htb]
\begin{minipage}[b]{1.0\linewidth}
  \centerline{\includegraphics[width=5.5cm]{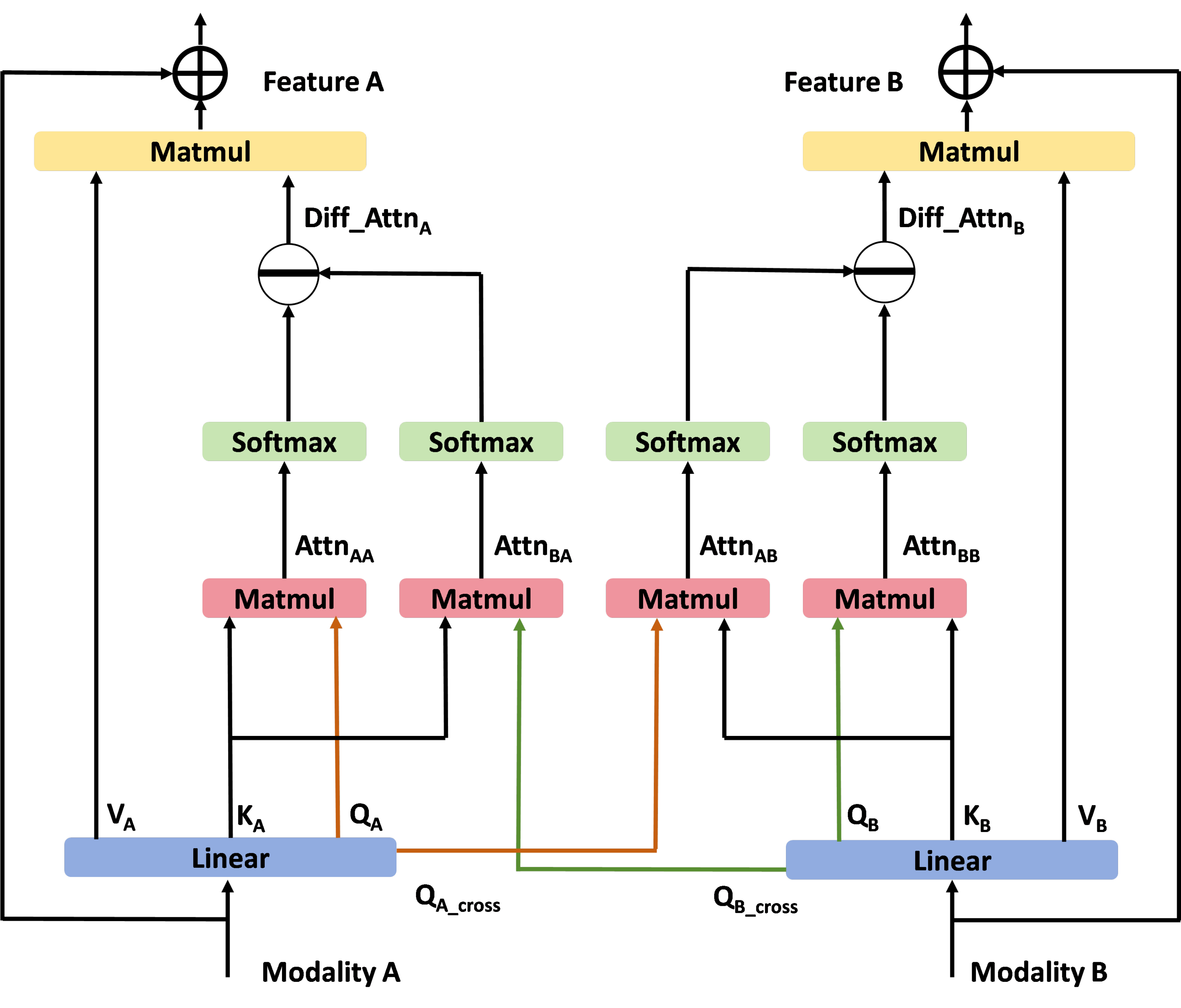}}
\end{minipage}
\caption{Differential cross-modal attention module}
\label{img3}
\vspace{-0.5cm}
\end{figure}

Inspired by the Differential Transformer \cite{i3}, we propose a differential cross-modal attention(DCA, \textbf{Fig.\ref{img3}}) that is more suitable for the multi-modal deepfake detection tasks. Taking the modality $A$ branch as an example, $Q_{B\_cross}$, $Q_A$, $K_A$, and $V_A$ derived through four linear layers. Following traditional attention mechanisms, the attention matrix is computed via matrix multiplication as follows:\\
\begin{equation}
\left\{
\begin{aligned}
Attn_{BA} &= Q_{B\_cross} K_A^{T} \\
Attn_{AA} &= Q_A K_A^{T}
\end{aligned}
\right.
\end{equation}

We define $Attn_{AA}-Attn_{BA}$ as the cross-modal attention matrix $Diff\_Attn_{A}$. Ultimately, the cross-modal output is obtained by multiplying $Diff\_Attn_{A}$ and $V_A$. 
The cross-modal alignment loss enforces that the cross-modal similarity of fake videos tends toward 0. Since $Attn_{AA}$ is a self-attention matrix, its intrinsic similarity remains unaffected by the cross-modal loss. In contrast, $Attn_{BA}$ functions as a cross-modal attention matrix: the lower the cross-modal similarity of a fake video, the stronger the constraints imposed on $Attn_{BA}$. This mechanism enhances $Diff\_Attn_{A}$ for fake videos, thereby facilitating more effective capture of forgery traces. The modality $B$ branch is the same as $A$.
\subsubsection{Multi-scale self-attention module}
\begin{figure}[htb]
\begin{minipage}[b]{1.0\linewidth}
  \centerline{\includegraphics[width=5.5cm]{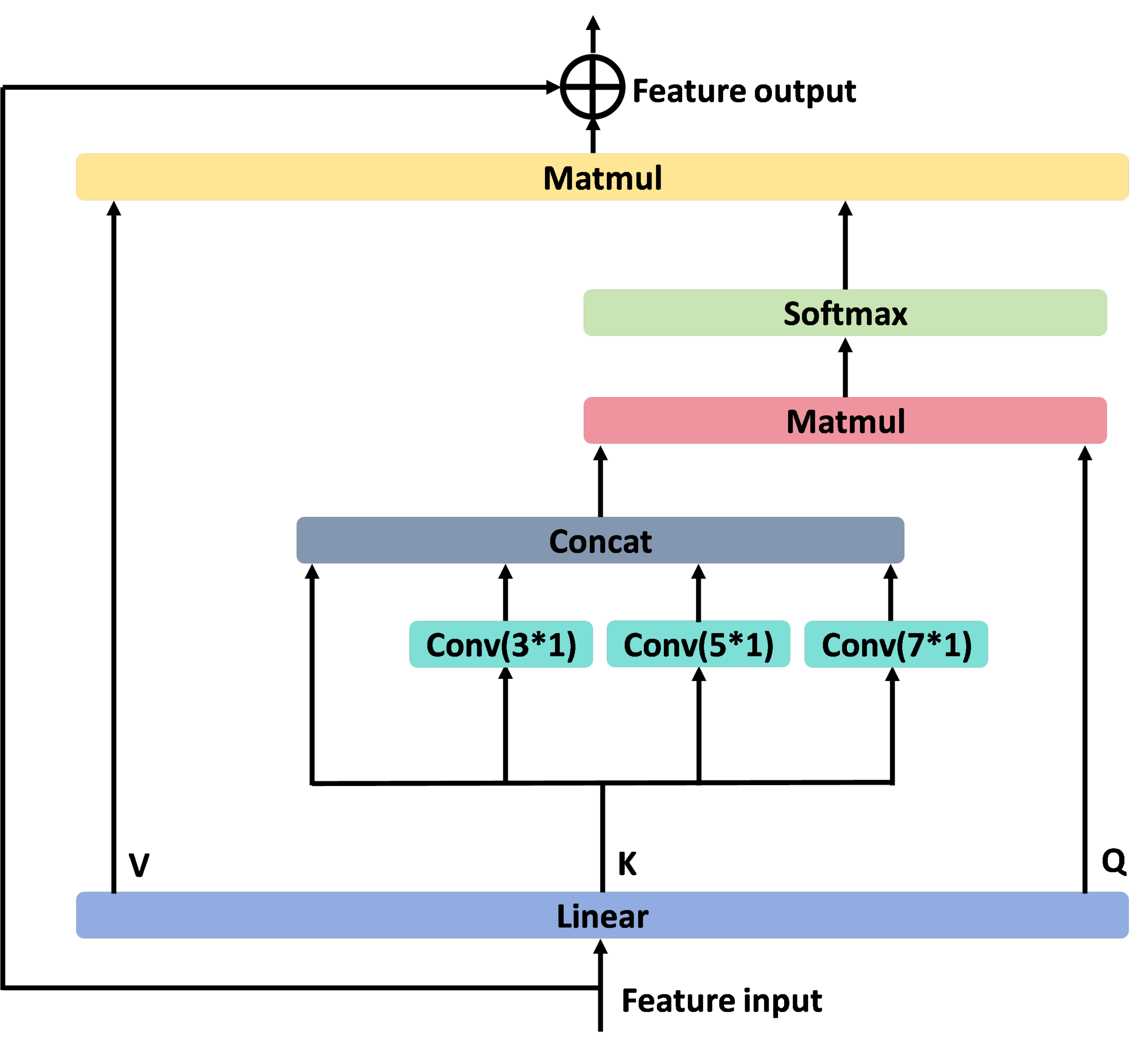}}
\end{minipage}
\caption{Multi-scale self-attention module}
\label{img4}
\vspace{-0.5cm}
\end{figure}

The traditional multi-head attention module lacks the capability to integrate multi-scale information across different embeddings. To address this limitation, we proposed a Multi-scale Self-Attention module (MSSA, \textbf{Fig.\ref{img4}}).

The input vector is passed through three linear layers to obtain $Q, K, V \in \mathbb{R}^{B \times T \times C}$. These are transformed into multi-head representations $Q, K, V \in \mathbb{R}^{B \times h \times T \times C_{head}}$, where $h$ denotes the number of heads. We split $K$ into four parts along the head dimension, each part is processed via 2D convolution at a specific scale, yielding outputs $K_{i} \in \mathbb{R}^{B \times \frac{h}{4} \times T \times C_{head}}$. We then concatenate all $K_{i}$ along the head dimension and compute the matrix product with $Q$ to generate the attention matrix $Attn \in \mathbb{R}^{B \times h \times T \times T}$. Finally, the output is obtained by multiplying $Attn$ with $V$.

The $K$ matrix is processed using 2D convolution, which integrates information from adjacent embeddings in a multi-scale manner. This operation enhances the representational capacity of embeddings and endows the transformer with greater flexibility in the time domain.
\subsection{Loss function}
We utilize cross-entropy loss ($L_{ce}$) to regularize single-modal classification and optimize multi-modal classification:
\begin{equation}
L_{ce}^{m}=-\sum_{c=1}^{k}(y_{m}^{c} \times log \frac{exp(f_{\theta}(x_{m})^{c})}{\sum_{c=1}^{k}exp(f_{\theta}(x_{m})^{c})})
\end{equation}

Where $k=2$ for binary deepfake detection, and $f_{\theta}(\cdot)$ represents the classification module. $x_{m}$ (where $m=[a,v,a\_v]$) serves as the input to the classification module. Additionally, we define the cross-modal alignment loss as follows:
\begin{equation}
\begin{aligned}
L_{c}=y_{a\_v}^{n} \times (1-d^n)+(1-y_{a\_v}^{n}) \times max(0,d^n)
\end{aligned}
\end{equation}

The similarity between modalities is measured by $d^n=\frac{x_a^n \cdot x_v^n}{\Vert x_a^n \Vert \Vert x_v^n \Vert}$. Where $n$ denotes the $n$-th sample in the batch, $x_a^n$ and $x_v^n$ denote the output embedding $[z_{1},..., z_{T}]$ from the transformer in each modal branch. Finally, we optimize the model using the following formula:
\begin{equation}
\begin{aligned}
L=\lambda_{ce}^{a} L_{ce}^{a}+\lambda_{ce}^{v} L_{ce}^{v}+\lambda_{ce}^{a\_v} L_{ce}^{a\_v}+\lambda_{c} L_{c}
\end{aligned}
\vspace{-0.4cm}
\end{equation}
\begin{table}[h!]
  \vspace{-0.3cm}
  \begin{center}
    \caption{Comparison with other methods on FakeAVCeleb}
    \label{table1}
    \vspace{0.2cm}
    \begin{tabular}{l|c|c} 
      \textbf{Method} & \textbf{ACC $\uparrow$} & \textbf{AUC $\uparrow$}\\
      \hline
      VFD \cite{a4}& 81.52& 86.11\\
      MDS \cite{i10}& 82.80& 86.50\\
      AVOID-DF \cite{a5}& 83.70& 89.20\\
      MRDF-CE \cite{i2}& 94.05& 92.43\\
      BusterX \cite{m2}& 96.30& -\\
      \hline
      \boldmath{Ours}& \textbf{98.75}& \textbf{98.83}\\
    \end{tabular}
  \end{center}
  \vspace{-0.5cm}
\end{table}
\section{EXPERIMENTS}
\subsection{Datasets}
We evaluated our method on the public dataset FakeAVCeleb \cite{i4}. This dataset comprises 500 real videos and over 20,000 fake videos, with data categorized into four distinct types: RealAudio-RealVideo (RARV), FakeAudio-RealVideo (FARV), RealAudio-FakeVideo (RAFV), and FakeAudio-FakeVideo (FAFV). For the sake of fairness in evaluation, we divided the dataset such that the four data types were maintained at a 1:1:1:1 ratio.
Given the significant redundancy present in video frames, we employed DLIB to locate the key facial regions in each frame. We then cropped these facial regions to serve as the core input frames for subsequent processing.
\subsection{Experimental setup}
Following the design paradigm of prior audio-visual methods \cite{i2}, we adopt a linear projection layer as the audio pre-encoder. To better capture discriminative information from video frames, we use an adapted Res2Net \cite{a7} as the visual pre-encoder---specifically, integrating a wavelet convolution module \cite{i5} at the Res2Net's front end to extract multi-scale visual features. Additionally, we introduce the Convolutional Block Attention Module (CBAM) \cite{i6} between consecutive Res2Net backbone layers, boosting the model's feature expression.
For multi-modal feature fusion and further processing, the audio-visual transformer module consists of 6 transformer blocks---this balances model capacity and computational efficiency. During training, the model was optimized with the Adam optimizer for 200 epochs to ensure convergence to stable performance.
\section{RESULTS AND ANALYSIS}
In this section, we evaluated the performance of the model. In addition, each module was analyzed in detail through ablation experiments.
\subsection{Test results and analysis}
We compare the performance of our proposed model against other methods on the FakeAVCeleb dataset, with detailed results presented in \textbf{Table \ref{table1}}. Notably, our model achieves competitive and strong performance on the FakeAVCeleb dataset: it reaches a classification accuracy of $98.75 \%$ and an AUC score of $98.83 \%$. These results explicitly demonstrate that our model exhibits strong discriminative ability in distinguishing between real and fake audio-visual content compared to existing advanced baselines.
\subsection{Ablation study}
To boost our model's fake audio-visual cue extraction, the transformer encoder adopts differential cross-modal attention ($DCA$) and multi-scale self-attention $(MSSA$). To compare the performance of each module, we set up four experiments by replacing the attention layer in the transformer. We assume that $CA$ and $SA$ represent traditional cross-modal attention and traditional self-attention, respectively. As shown in \textbf{Table \ref{table2}}, both $DCA$ and $MSSA$ improve model performance, with $DCA$ bringing a more pronounced gain.
\begin{table}[h!]
  \vspace{-0.4cm}
  \begin{center}
    \caption{Ablation study}
    \label{table2}
    \vspace{0.2cm}
    \begin{tabular}{l|c|c} 
      \textbf{Model} & \textbf{ACC $\uparrow$} & \textbf{AUC $\uparrow$}\\
      \hline
      $CA + SA$& 96.75& 96.17\\
      $CA + MSSA$& 97.00& 97.00\\
      $DCA + SA$& 98.00& 98.00\\
      \boldmath{$DCA + MSSA$}& \textbf{98.75}& \textbf{98.83}\\
    \end{tabular}
  \end{center}
  \vspace{-0.5cm}
\end{table}
\subsection{Visual Results and Analysis}
\begin{figure}[htb]
\begin{minipage}[b]{.48\linewidth}
  \centering
  \centerline{\includegraphics[width=4.5cm]{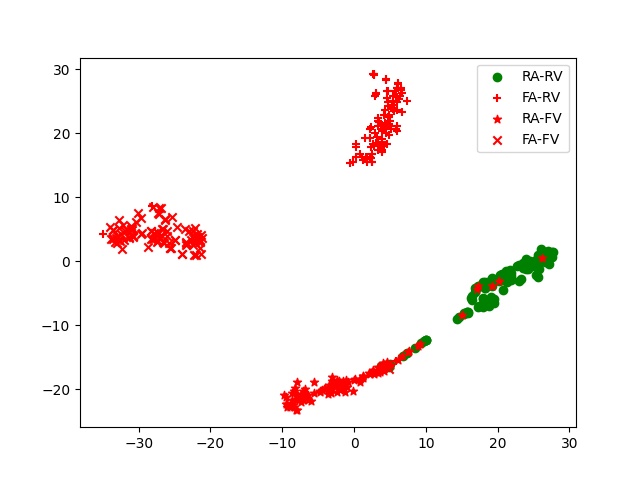}}
  \centerline{(a) $CA+SA$}
\end{minipage}
\hfill
\begin{minipage}[b]{0.48\linewidth}
  \centering
  \centerline{\includegraphics[width=4.5cm]{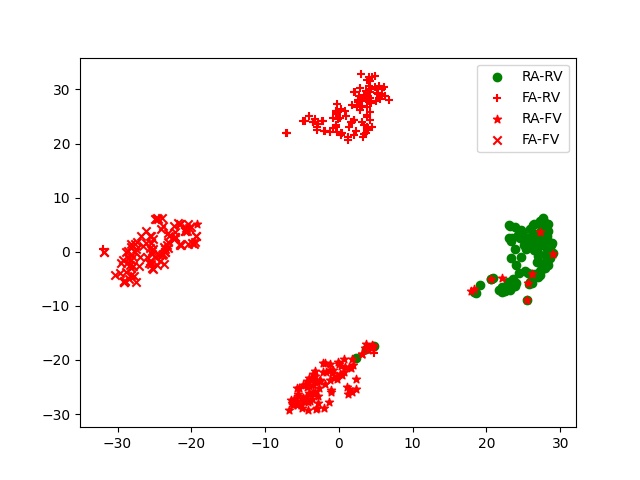}}
  \centerline{(b) $CA+MSSA$}
\end{minipage}
\begin{minipage}[b]{.48\linewidth}
  \centering
  \centerline{\includegraphics[width=4.5cm]{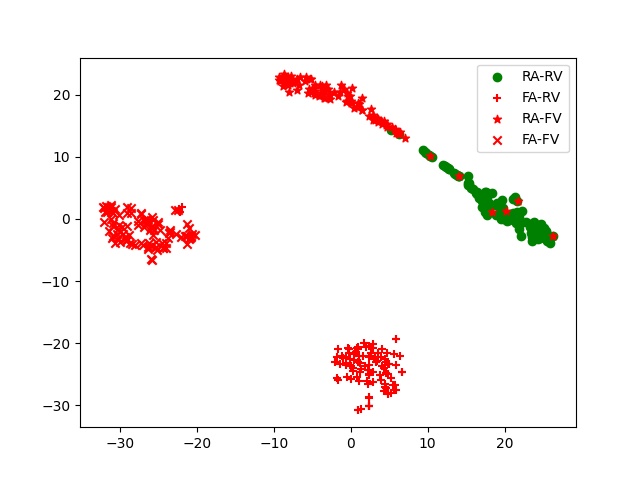}}
  \centerline{(c) $DCA+SA$}
\end{minipage}
\hfill
\begin{minipage}[b]{0.48\linewidth}
  \centering
  \centerline{\includegraphics[width=4.5cm]{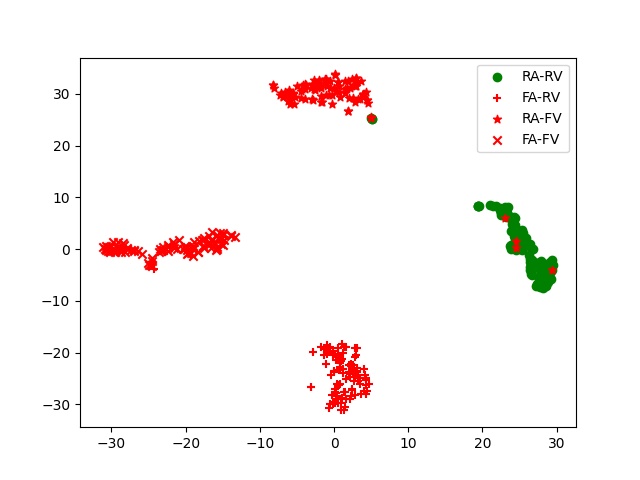}}
  \centerline{(d) $DCA+MSSA$}
\end{minipage}
\caption{Comparison of T-SNE results of different models.}
\label{img5}
\vspace{-0.5cm}
\end{figure}

We visualized the deepfake prediction results of our model and baseline methods via T-SNE in \textbf{Fig.\ref{img5}}. As clearly illustrated in the figure, the baseline model struggles to distinguish between the RA-RV and RA-FV categories, whereas our model---by expanding the embedding perception scale and replacing cross-modal attention---achieves enhanced discriminative ability.
\section{CONCLUSION}
This paper proposes two novel attention modules specifically designed for integration into the transformer encoder of multi-modal deepfake detection systems. Specifically, cross-modal differential attention enhances the model's compatibility with multi-modal deepfake detection tasks by leveraging attention matrix differences. Multi-scale self-attention boosts the model's ability to perceive adjacent embeddings via multiple convolutional layers, enabling multi-scale visual perception. Compared with representative existing methods, our approach achieves competitive performance on the public dataset.
\vfill\pagebreak
\bibliographystyle{IEEEbib}
\bibliography{strings}

@article{a1,
  title={Auto-Encoding Variational Bayes},
  author={ Kingma, Diederik P  and  Welling, Max },
  journal={arXiv.org},
  year={2014},
}

@article{a2,
author = {Goodfellow, Ian and Pouget-Abadie, Jean and Mirza, Mehdi and Xu, Bing and Warde-Farley, David and Ozair, Sherjil and Courville, Aaron and Bengio, Yoshua},
title = {Generative adversarial networks},
year = {2020},
issue_date = {November 2020},
publisher = {Association for Computing Machinery},
address = {New York, NY, USA},
volume = {63},
number = {11},
issn = {0001-0782},
url = {https://doi.org/10.1145/3422622},
doi = {10.1145/3422622},
journal = {Commun. ACM},
month = oct,
pages = {139–144},
numpages = {6}
}

@ARTICLE{a3,
  author={Liu, Miao and Wang, Jing and Qian, Xinyuan and Li, Haizhou},
  journal={IEEE Transactions on Circuits and Systems for Video Technology}, 
  title={Audio-Visual Temporal Forgery Detection Using Embedding-Level Fusion and Multi-Dimensional Contrastive Loss}, 
  year={2024},
  volume={34},
  number={8},
  pages={6937-6948},
  doi={10.1109/TCSVT.2023.3326694}}

@article{a4,
author = {Cheng, Harry and Guo, Yangyang and Wang, Tianyi and Li, Qi and Chang, Xiaojun and Nie, Liqiang},
title = {Voice-Face Homogeneity Tells Deepfake},
year = {2023},
issue_date = {March 2024},
publisher = {Association for Computing Machinery},
address = {New York, NY, USA},
volume = {20},
number = {3},
issn = {1551-6857},
url = {https://doi.org/10.1145/3625231},
doi = {10.1145/3625231},
journal = {ACM Trans. Multimedia Comput. Commun. Appl.},
month = nov,
articleno = {76},
numpages = {22}
}

@ARTICLE{a5,
  author={Yang, Wenyuan and Zhou, Xiaoyu and Chen, Zhikai and Guo, Bofei and Ba, Zhongjie and Xia, Zhihua and Cao, Xiaochun and Ren, Kui},
  journal={IEEE Transactions on Information Forensics and Security}, 
  title={AVoiD-DF: Audio-Visual Joint Learning for Detecting Deepfake}, 
  year={2023},
  volume={18},
  number={},
  pages={2015-2029},
  doi={10.1109/TIFS.2023.3262148}}

@article{a7,
author = {Gao, Shang-Hua and Cheng, Ming-Ming and Zhao, Kai and Zhang, Xin-Yu and Yang, Ming-Hsuan and Torr, Philip},
title = {Res2Net: A New Multi-Scale Backbone Architecture},
year = {2021},
issue_date = {Feb. 2021},
publisher = {IEEE Computer Society},
address = {USA},
volume = {43},
number = {2},
issn = {0162-8828},
url = {https://doi.org/10.1109/TPAMI.2019.2938758},
doi = {10.1109/TPAMI.2019.2938758},
journal = {IEEE Trans. Pattern Anal. Mach. Intell.},
month = feb,
pages = {652–662},
numpages = {11}
}

@inproceedings{i1,
 author = {Ho, Jonathan and Jain, Ajay and Abbeel, Pieter},
 booktitle = {Advances in Neural Information Processing Systems},
 editor = {H. Larochelle and M. Ranzato and R. Hadsell and M.F. Balcan and H. Lin},
 pages = {6840--6851},
 publisher = {Curran Associates, Inc.},
 title = {Denoising Diffusion Probabilistic Models},
 url = {https://proceedings.neurips.cc/paper_files/paper/2020/file/4c5bcfec8584af0d967f1ab10179ca4b-Paper.pdf},
 volume = {33},
 year = {2020}
}

@INPROCEEDINGS{i2,
  author={Zou, Heqing and Shen, Meng and Hu, Yuchen and Chen, Chen and Chng, Eng Siong and Rajan, Deepu},
  booktitle={ICASSP 2024 - 2024 IEEE International Conference on Acoustics, Speech and Signal Processing (ICASSP)}, 
  title={Cross-Modality and Within-Modality Regularization for Audio-Visual Deepfake Detection}, 
  year={2024},
  volume={},
  number={},
  pages={4900-4904},
  doi={10.1109/ICASSP48485.2024.10447248}}

@inproceedings{i3,
title={Differential Transformer},
author={Tianzhu Ye and Li Dong and Yuqing Xia and Yutao Sun and Yi Zhu and Gao Huang and Furu Wei},
booktitle={The Thirteenth International Conference on Learning Representations},
year={2025},
url={https://openreview.net/forum?id=OvoCm1gGhN}
}

@inproceedings{i4,
title={Fake{AVC}eleb: A Novel Audio-Video Multimodal Deepfake Dataset},
author={Hasam Khalid and Shahroz Tariq and Minha Kim and Simon S. Woo},
booktitle={Thirty-fifth Conference on Neural Information Processing Systems Datasets and Benchmarks Track (Round 2)},
year={2021},
url={https://openreview.net/forum?id=TAXFsg6ZaOl}
}

@inproceedings{i5,
author = {Finder, Shahaf E. and Amoyal, Roy and Treister, Eran and Freifeld, Oren},
title = {Wavelet Convolutions for Large Receptive Fields},
year = {2024},
isbn = {978-3-031-72948-5},
publisher = {Springer-Verlag},
address = {Berlin, Heidelberg},
url = {https://doi.org/10.1007/978-3-031-72949-2_21},
doi = {10.1007/978-3-031-72949-2_21},
booktitle = {Computer Vision – ECCV 2024: 18th European Conference, Milan, Italy, September 29–October 4, 2024, Proceedings, Part LIV},
pages = {363–380},
numpages = {18},
location = {Milan, Italy}
}

@inproceedings{i6,
author = {Woo, Sanghyun and Park, Jongchan and Lee, Joon-Young and Kweon, In So},
title = {CBAM: Convolutional Block Attention Module},
year = {2018},
isbn = {978-3-030-01233-5},
publisher = {Springer-Verlag},
address = {Berlin, Heidelberg},
url = {https://doi.org/10.1007/978-3-030-01234-2_1},
doi = {10.1007/978-3-030-01234-2_1},
booktitle = {Computer Vision – ECCV 2018: 15th European Conference, Munich, Germany, September 8–14, 2018, Proceedings, Part VII},
pages = {3–19},
numpages = {17},
location = {Munich, Germany}
}

@INPROCEEDINGS{i7,
  author={Chen, Xiaohuan and Lu, Wenhuan and Zhang, Ruiteng and Xu, Junhai and Lu, Xugang and Zhang, Lin and Wei, Jianguo},
  booktitle={ICASSP 2025 - 2025 IEEE International Conference on Acoustics, Speech and Signal Processing (ICASSP)}, 
  title={Continual Unsupervised Domain Adaptation for Audio Deepfake Detection}, 
  year={2025},
  volume={},
  number={},
  pages={1-5},
  doi={10.1109/ICASSP49660.2025.10890538}}

@INPROCEEDINGS{i8,
  author={Xu, Yuting and Liang, Jian and Jia, Gengyun and Yang, Ziming and Zhang, Yanhao and He, Ran},
  booktitle={2023 IEEE/CVF International Conference on Computer Vision (ICCV)}, 
  title={TALL: Thumbnail Layout for Deepfake Video Detection}, 
  year={2023},
  volume={},
  number={},
  pages={22601-22611},
  doi={10.1109/ICCV51070.2023.02071}}

@inproceedings{i9,
author = {Mittal, Trisha and Bhattacharya, Uttaran and Chandra, Rohan and Bera, Aniket and Manocha, Dinesh},
title = {Emotions Don't Lie: An Audio-Visual Deepfake Detection Method using Affective Cues},
year = {2020},
isbn = {9781450379885},
publisher = {Association for Computing Machinery},
address = {New York, NY, USA},
url = {https://doi.org/10.1145/3394171.3413570},
doi = {10.1145/3394171.3413570},
booktitle = {Proceedings of the 28th ACM International Conference on Multimedia},
pages = {2823–2832},
numpages = {10},
location = {Seattle, WA, USA},
series = {MM '20}
}

@inproceedings{i10,
author = {Chugh, Komal and Gupta, Parul and Dhall, Abhinav and Subramanian, Ramanathan},
title = {Not made for each other- Audio-Visual Dissonance-based Deepfake Detection and Localization},
year = {2020},
isbn = {9781450379885},
publisher = {Association for Computing Machinery},
address = {New York, NY, USA},
url = {https://doi.org/10.1145/3394171.3413700},
doi = {10.1145/3394171.3413700},
booktitle = {Proceedings of the 28th ACM International Conference on Multimedia},
pages = {439–447},
numpages = {9},
location = {Seattle, WA, USA},
series = {MM '20}
}

@inproceedings{i12, series={MM ’21},
   title={Is Someone Speaking?: Exploring Long-term Temporal Features for Audio-visual Active Speaker Detection},
   url={http://dx.doi.org/10.1145/3474085.3475587},
   DOI={10.1145/3474085.3475587},
   booktitle={Proceedings of the 29th ACM International Conference on Multimedia},
   publisher={ACM},
   author={Tao, Ruijie and Pan, Zexu and Das, Rohan Kumar and Qian, Xinyuan and Shou, Mike Zheng and Li, Haizhou},
   year={2021},
   month=oct, pages={3927–3935},
   collection={MM ’21} }

@misc{m2,
      title={BusterX: MLLM-Powered AI-Generated Video Forgery Detection and Explanation}, 
      author={Haiquan Wen and Yiwei He and Zhenglin Huang and Tianxiao Li and Zihan Yu and Xingru Huang and Lu Qi and Baoyuan Wu and Xiangtai Li and Guangliang Cheng},
      year={2025},
      eprint={2505.12620},
      archivePrefix={arXiv},
      primaryClass={cs.CV},
      url={https://arxiv.org/abs/2505.12620}, 
}

\end{document}